\documentclass[sigconf,screen,nonacm]{acmart}

\renewcommand\footnotetextcopyrightpermission[1]{}
\usepackage[T1]{fontenc}
\usepackage[utf8]{inputenc}
\usepackage{microtype}
\usepackage{amsmath}

\usepackage{amssymb}
\usepackage{booktabs}
\usepackage{multirow}
\usepackage{xcolor}
\usepackage{graphicx}

\DeclareUnicodeCharacter{2018}{`}
\DeclareUnicodeCharacter{2019}{'}
\DeclareUnicodeCharacter{201C}{``}
\DeclareUnicodeCharacter{201D}{''}
\DeclareUnicodeCharacter{2013}{--}
\DeclareUnicodeCharacter{2014}{---}

\def\eg{\emph{e.g.}}

\title{Explaining and Tuning Transformer-based LLMs in Arithmetic Tasks with Human Strategies}

\author{%
{Luyu QIU}\textsuperscript{1},
{Jianing LI}\textsuperscript{2},
{Hwanhee KIM}\textsuperscript{3},
{Xiaoyong WEI}\textsuperscript{2},
{Yueyuan ZHENG}\textsuperscript{1},
{Janet HSIAO}\textsuperscript{1},
{Lei CHEN}\textsuperscript{1}}
\affiliation{%
  \institution{\textsuperscript{1}The Hong Kong University of Science and Technology\\
  \textsuperscript{2}The Hong Kong Polytechnic University\\
  \textsuperscript{3}University of California, Berkeley}
  \country{}
}

\setcopyright{none}
\acmDOI{}
\acmISBN{}

\begin{document}

\begin{abstract}
Transformer-based large language models (LLMs) continue to achieve state-of-the-art performance across various natural language processing tasks. However, their subpar performance on seemingly elementary problems, such as basic arithmetic, raises concerns about model reliability, safety, and ethical deployment. In this study, we demonstrate that the performance of a vanilla Transformer model trained on integer arithmetic tasks can be improved using methods effective for human learners. We begin by decomposing the arithmetic task into well-defined subtasks and conducting loss convergence order analysis together with ablation studies for each subtask. Our findings reveal that LLMs exhibit learning patterns similar to those of human learners, with a faster learning speed for simpler subtasks compared to more complex ones. In addition, we successfully improved the accuracy of LLMs by applying problem-solving strategies and cognitive empowerment methods shown to enhance the performance of human learners. This suggests that transformer-based LLMs may share cognitive processes with human learners in arithmetic. Lastly, we provide a comprehensive demonstration of our method's effectiveness, including significant accuracy improvement experiments, visualization verification, and explanation-based analysis to illuminate the intricacies of LLMs in arithmetic learning. In general, this work explores the potential similarities between transformer-based LLMs and human learners, supported by explainable AI (XAI) verifications, ultimately fostering trust in LLMs for critical and high-stakes applications.
\end{abstract}


\maketitle
\pagestyle{plain}
\section{Introduction}\label{introduction}
%% Labels are used to cross-reference an item using \ref command.

In recent years, Natural Language Processing (NLP) has witnessed remarkable progress, largely driven by the advent of transformer architectures~\cite{vaswani2017attention}, which form the backbone of modern Large Language Models (LLMs). These models have demonstrated remarkable performance on a wide array of tasks, such as dialogue, conversational agents, and machine translation~\cite{devlin2018bert,raffel2020exploring,touvron2023llama,dziri2024faith}. The emergence of models such as GPT-4~\cite{achiam2023gpt} has further sparked widespread enthusiasm by pushing the boundaries of Artificial General Intelligence (AGI).

Despite these notable achievements, LLMs exhibit striking weaknesses in seemingly elementary arithmetic tasks, \eg, the modern LLM GPT-4~\cite{achiam2023gpt} even struggles with tasks such as basic integer calculations~\cite{dziri2024faith}, as evidenced by the accuracy reported in Table~\ref{table:gpt4acc}. This stark contrast between LLMs’ advanced capabilities in complex tasks and their shortcomings in trivial arithmetic not only raises concerns regarding reliable and ethical deployment, but also prompts fundamental questions about the true scope and limitations of current LLM technology.

%算术任务和NLP任务的区别，论证在算术任务上研究LLMs是有意义的
Arithmetic tasks differ from typical NLP tasks in many aspects, with the most prominent distinction being the uniqueness of the results. Although NLP tasks often allow multiple valid outputs and encourage diversity, arithmetic problems generally have only one definitive answer. In addition, different parts of the input vary in importance in NLP tasks, and omitting some less important elements may not affect the outcome. However, every digit is crucial in arithmetic tasks to ensure accuracy. Moreover, arithmetic calculations involve multiple complex intermediate steps, so precisely handling these steps is critical for transformer-based models. Finally, unlike natural left-to-right processing in most NLP tasks, arithmetic operations such as addition or multiplication often benefit from starting with the least significant digit. Due to these special characteristics of arithmetic tasks outlined above, exploring explainable approaches for Transformer-based LLMs in this domain is highly valuable, as the distinctive nature of arithmetic tasks provides an excellent opportunity to understand and uncover the internal mechanisms of Transformer-based models.

\begin{table}
\begin{center}
\caption{The accuracy (\%) of GPT-4 on 1-5 digits arithmetic tasks. We use the prompt “What's the answer of $a\bigodot b$?”, where $\bigodot$ denotes the four operation symbols, \eg, $+-\times \div$. Each combination of operator and digit is tested 100 times.}
\setlength{\tabcolsep}{10 pt}
\small
\begin{tabular}{c|ccccc}
\toprule
\multirow{2}{*}{Type} &\multicolumn{5}{c}{\# Digits}\\\cmidrule{2-6}
          &(1,1) &(2,2) &(3,3) &(4,4) &(5,5) \\\midrule
+         &100 &100 &100 &99  &100 \\\midrule
-         &100 &100 &99  &96  &99  \\\midrule
$\times$  &99  &82  &43  &13  &5  \\\midrule
$\div$    &94  &58  &29  &6   &0   \\\bottomrule
\end{tabular}
\label{table:gpt4acc}
\end{center}
\end{table}

%现有相关工作的不足
Existing eXplainable AI (XAI) methods for Transformers predominantly focus on interpretability in specific instances of certain tasks, such as indirect object identification \cite{wang2022interpretability} and colored objects \cite{merullo2023circuit}. While these methods provide valuable insights into how Transformers make decisions, they often fail to capture the intrinsic correlations in complex tasks or the broader decision-making process. Furthermore, the inherent complexity of multilayered self-attention architectures poses significant challenges in explaining how transformers produce specific outputs \cite{vig2019multiscale}. Providing explanations for Transformers on arithmetic tasks remains an open research challenge, with only a limited amount of work \cite{dziri2024faith,lee2023teaching} addressing this area. However, existing approaches largely treat these models as black boxes, offering little insight into their internal mechanisms. 

%本文
In this paper, we uncovered the learning process of transformer-based LLMs in integer arithmetic tasks through problem analysis, and examined whether strategies and capacities critical to human arithmetic learning are effective in improving LLM's performance. Specifically, we reformulate traditional arithmetic rules into a Transformer-friendly framework. We adopted subtask decomposition methods, learning curve analysis, and model visualization analysis to diagnose accuracy issues and uncover their root causes. We then integrate human arithmetic strategies and enhance relevant capacities to gain a deeper understanding of the model's limitations and potential. Our findings provide valuable insight into understanding and interpreting Transformer-based models. The main contributions of this work are threefold:
\begin{itemize}
\item We propose a comprehensive problem analysis framework to explain how Transformer-based LLMs handle arithmetic tasks.
\item Based on our interpretive analysis, we enhance the model’s arithmetic performance and clarify the principles behind these improvements.
\item Our approach contributes to broader research on model understanding and interpretability, opening up possibilities for analyzing more complex tasks and Transformer architectures. In addition, these insights have wider implications for AI safety and reliability.
\end{itemize}

\begin{figure}
\centering
\includegraphics[width=0.8\linewidth]{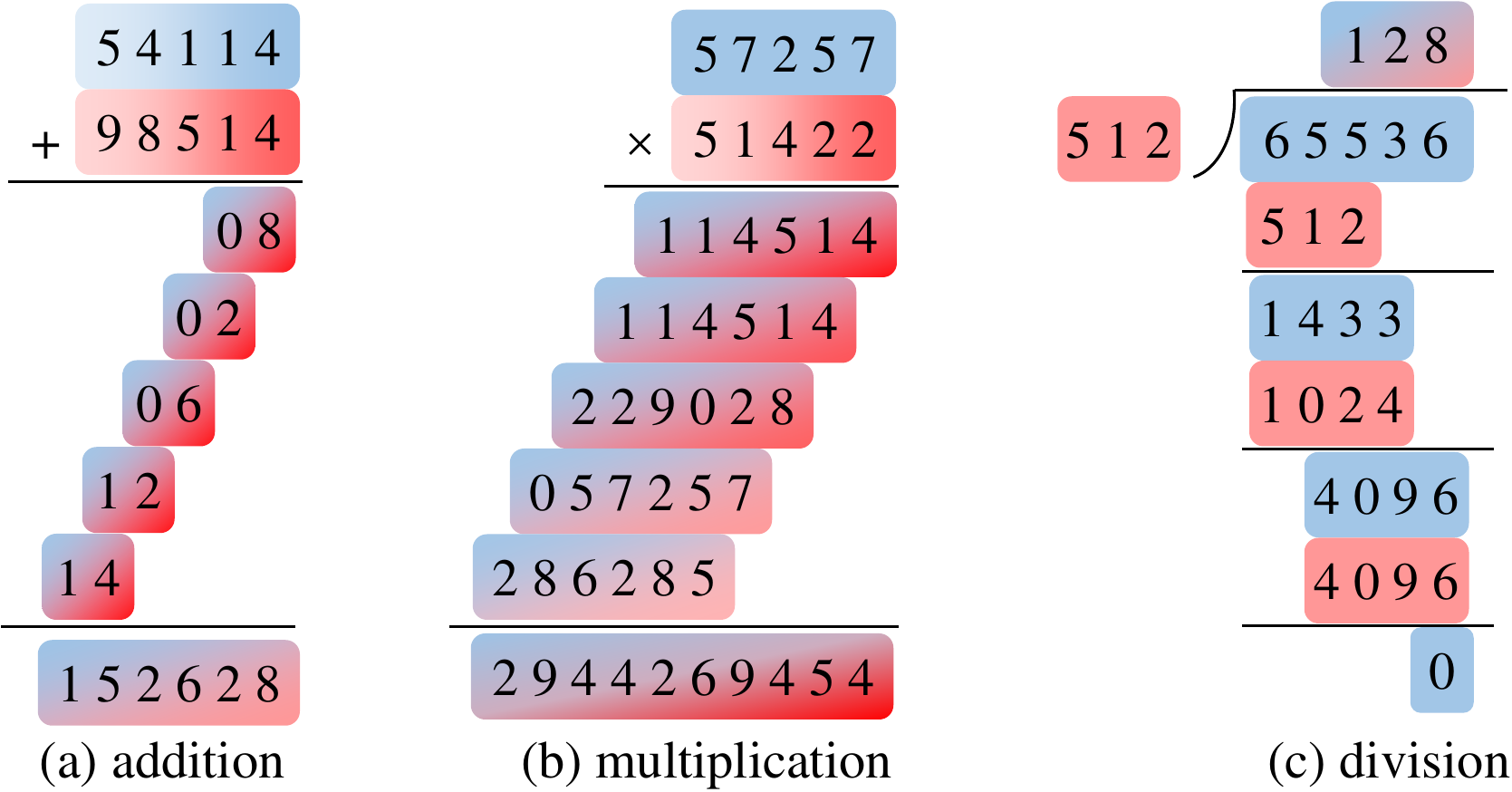}\\
\caption{The decomposed steps of typical arithmetic tasks, include (a) addition, (b) multiplication, and (c) division.}
% \vspace{-1ex}
\label{fig:div}
\end{figure}

\section{Related Work}
Since this work focuses on XAI techniques for transformer-based models in arithmetic tasks, the related work will address transformers in arithmetic tasks and their XAI methods.

\subsection{Transformers in Arithmetic Tasks}
Transformers~\cite{vaswani2017attention} use the self-attention mechanism to capture long-range dependencies among tokens and have been widely applied in both natural language processing~\cite{devlin2018bert,taylor2022galactica,thoppilan2022lamda,chung2024scaling} and computer vision~\cite{radford2021learning,li2023blip,li2024mini}.
Recent studies have explored the application of transformers in arithmetic tasks. ~\cite{qian2022limitations} identify the limitations of language models in arithmetic tasks and propose to introduce fine-grained computation steps to mitigate the shortcomings of transformer.
~\cite{yang2023gpt} focus on enhancing the reasoning process by teaching the model step by step. They developed MathGLM, a 2B parameter model that outperforms GPT-4 on several arithmetic problems, although its performance remains imperfect.
~\cite{lee2023teaching} investigate the impact of data format design in arithmetic tasks, finding that the use of detailed, instructive data with intermediate steps improves both accuracy and sample complexity.
~\cite{dziri2024faith} explore Transformers' limitations in compositional tasks like multi-digit multiplication, formulating these tasks as computation graphs to systematically measure complexity. They find that Transformer's performance deteriorates rapidly as task difficulty increases. 
These methods analyze the transformer's shortcomings in arithmetic tasks primarily through accuracy, offering limited insight into the internal mechanisms into how they perform computations. Moreover, the proposed multi-step decomposition refinements significantly reduce training and inference efficiency.

\subsection{XAI in Transformers}
Explainable AI (XAI) refers to a set of techniques and methods that make the decisions and outputs of AI systems understandable and interpretable for humans. Early XAI methods utilize the gradients of the output with respect to the input features to understand and interpret the model's predictions~\cite{zhou2016learning,selvaraju2017grad}.
Recently, numerous XAI studies have sought to explain how Transformers make decisions in a variety of tasks~\cite{wang2022interpretability,merullo2023circuit,quirke2023understanding}.
~\cite{wang2022interpretability} identifies the function of different attention heads and groups them into seven primary classes for an indirect-object identification task. 
~\cite{merullo2023circuit} propose to adjust the attention heads in the middle layers to ‘repair’ the Transformer's prediction in the Colored Objects task.
~\cite{quirke2023understanding} conducted a comprehensive analysis of a one-layer Transformer on n-digit integer addition by decomposing the task into three base functions and examining the learning conditions of each function. 
~\cite{shen2023positional} shows that reliance on positional information causes poor performance on arithmetic problems involving fewer digits, and proposes corresponding refinements.
Although existing studies have offered some level of interpretability for Transformer-based models and provided insights that improve accuracy, they are either not conducted on arithmetic tasks or lack in-depth analysis of how models perform tasks, lacking a theoretical foundation for a more systematic explainability approach.

\textbf{Comparison with existing methods:}
There are some works focused on arithmetic tasks in transformers~\cite{dziri2024faith,lee2023teaching}. However, these studies differ significantly from ours in several ways:
1) While these studies are data-driven and treat Transformers as black boxes, focusing on data variations to evaluate capacity and limitations. Differently, our work instead interprets the internal mechanisms of the model.
2) ~\cite{dziri2024faith} decomposes arithmetic into sequential subtasks using complex prompts that reflect human reasoning. In contrast, we use standard arithmetic equations and analyze how specific attention heads handle these subtasks.
3) ~\cite{lee2023teaching} explores how data formats affect task accuracy, whereas we examine their impact on both accuracy and the internal learning processes of the model.

Despite these differences, our findings are consistent with previous work. For instance, ~\cite{dziri2024faith} decomposes arithmetic into subtasks, while ~\cite{lee2023teaching} shows how reversed outputs help the model focus on fewer digits. This paper provides a detailed analysis of the model's computation process, visually demonstrating how Transformers handle arithmetic tasks and proposing improvements to address their limitations. Our work and ~\cite{dziri2024faith} offer complementary perspectives on Transformer interpretability.

\section{Methodology}
\subsection{Comprehensive problem analysis}

% \begin{figure}
% \centering
% \includegraphics[width=0.8\linewidth]{img/framework.png}\\
% % \vspace{-1ex}
% \caption{Cognition-driven XAI Framework (CDXF)}
% % \vspace{-2ex}
% \label{fig:framework}
% \end{figure}

We propose a comprehensive problem analysis framework to inspect the learning process of the models, which enhances the interpretability of the model behavior with theoretical support from cognitive science. The framework has three steps: 1) subtask decomposition, 2) learning curve analysis, and 3) model visualization analysis. In human problem solving, human learners often break down complex tasks into a combination of simpler tasks \cite{lee2001does}, and they start from easier tasks before they can solve harder ones \cite{clements2020learning}, with different brain regions engaged in different subtasks \cite{arsalidou2018brain}. Here we decomposed the arithmetic tasks into smaller components and examined whether LLMs performed subtasks differentially. We referred to this comparison method as the subtask decomposition analysis. This method helps to identify the subtasks that are particularly difficult for the LLMs. For each subtask, we inspected the learning curves for the output digits to examine the learning process for the specific calculation. We referred to this inspection method as the learning analysis method. This method facilitates the investigation of difficulties faced by LLMs in generating specific output digits. After inspecting the learning curve, we visualized the contribution of attention heads to the model output digit. The visualization method helps to examine whether the model follows humans' specialized functioning for different subtasks. 
\subsection{Human strategy and capacity transfer}
 Researchers in educational psychology and cognitive science have explored various learning and problem-solving strategies to improve the arithmetic performance of human learners. Here we tested two representative strategies, including computational strategies \cite{hiebert2013conceptual} and CoT reasoning \cite{lockwood2016algorithmic}, on transformer-based LLMs in solving arithmetic problems. In addition, previous research highlighted the importance of cognitive capacities, such as working memory capacity \cite{zhang2022relationship} and general intelligence \cite{Bor2010math}, in math learning. Therefore, we hypothesized that employing human computational strategy, adopting CoT prompting, and enhancing computational capacity could significantly improve model arithmetic performance. Indeed, previous research has demonstrated the effectiveness of CoT prompting \cite{imani2023mathprompter} in improving LLM's mathematical performance.

\subsection{Arithmetic Tasks Formulation}
This paper focuses on the integer arithmetic task in Transformer models. The input is a sequence of symbols, which consists of two $n$-digit operand $D=(D_{n-1},...,D_1,D_0)$ and $D'=(D_{n-1}',...,D_1',D_0')$, along with operators $\bigodot$, where $\bigodot$ can be $+$, $\times$ or $\div$. We did not study subtraction due to its similarity to addition.
Transformer first converts the input into a sequence of one-hot vectors representing corresponding symbols through vocabulary table of size $V$. These one-hot vectors are then mapped to a sequence of embeddings, $\textbf{x}=(x_1, x_2, ..., x_L)$, where $x_i\in \mathcal{R}^d$ is the embedding of $i$-th word with dimension $d$, $L$ is sequence length. After ``$=$", the model predicts the answer digits $A=(A_{m-1},...,A_1,A_0)$. 
An example of addition formula is shown as,
\begin{figure}[h]
\centering
% \vspace{-1ex}
\includegraphics[width=0.8\linewidth]{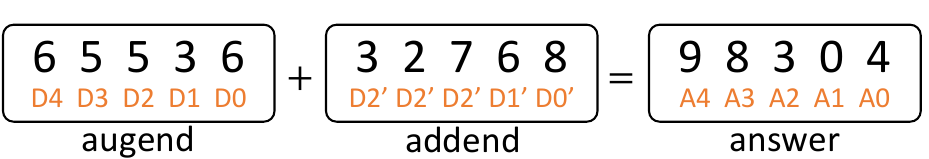}\\
% \vspace{-1ex}
\caption{An example of input addition formula.}
% \vspace{-2ex}
\label{fig:div_example}
\end{figure}

\begin{table}%[t]
\begin{center}
\caption{Examples of data format of arithmetic tasks. The symbols in red after ``$=$" are what the model needs to predict.}
\setlength{\tabcolsep}{10 pt}
\small
\begin{tabular}{l|l|l}
\toprule
Type &data      &Example  \\\midrule
addition &$m + m$    &$65536 + 65535 =\textcolor{red}{131071}$  \\\midrule

\multirow{9}{*}{multiply} 
&$m\times u$ (O)  &$57257\times 2=\textcolor{red}{114514}$      \\\cmidrule(r){2-3}
&$m\times u$ (R) &$57257\times 2=\textcolor{red}{415411}$      \\\cmidrule(r){2-3}
&$m\times m$ (O)  &$57257\times 51422=\textcolor{red}{2944269454}$  \\\cmidrule(r){2-3}
&$m\times m$ (R) &$57257\times 51422=\textcolor{red}{4549624492}$  \\\cmidrule(r){2-3}
&\multirow{6}{*}{$m\times m$ with CoT }      &$57257\times 5=\textcolor{red}{286285}$ \\
                    &&$57257\times 1=\ \textcolor{red}{57257} $ \\
                    &&$57257\times 4=\textcolor{red}{229028}$ \\
                    &&$57257\times 2=\textcolor{red}{114514}$ \\
                    &&$57257\times 2=\textcolor{red}{114514}$ \\
                    &&$57257\times 51422=\textcolor{red}{4549624492}$\\\midrule

\multirow{9}{*}{division} 
     &$m\div u$    &$131072\div 4 =\textcolor{red}{32768}$  \\\cmidrule(r){2-3}
     &$m\div m$    &$16777216\div 256 =\textcolor{red}{65536}$      \\\cmidrule(r){2-3}
     & \multirow{7}{*}{$m\div m$ with CoT}                        
                    &\ \ $167\div 256 =\textcolor{red}{0, 167} $ \\
                    &&$1677\div 256=\textcolor{red}{6, 141}$ \\
                    &&$1417\div 256=\textcolor{red}{5, 137}$ \\
                    &&$1372\div 256=\textcolor{red}{5, 092}$ \\
                    &&$0921\div 256=\textcolor{red}{3, 153}$ \\
                    &&$1536\div 256=\textcolor{red}{6, 000}$ \\
                    &&$16777216\div 256=\textcolor{red}{65536}$\\\midrule 
\end{tabular}
\label{table:data_format}
% \vspace{-20mm}
\end{center}
\end{table}

Accurate calculation of answer digits requires complex calculations. Following subtask decomposition observed in human behavior and considering the characteristics of the self-attention mechanism, we model arithmetic tasks as a combination of simple operations acting on digit pairs,
\begin{equation}
\begin{aligned}
a_i = \sum_{i=0}^n\sum_{j=0}^n f_{ij}(D_i, D_j').
\end{aligned}
\end{equation}
This data movement between tokens aligns well with the self-attention mechanism. 
Transformers need to learn distinct functions for different pairs of digits to perform arithmetic calculations. %,, which we will analyze in detail in the following sections.
In the following sections, we analyze and explain the transformer's learning process and implement for arithmetic tasks via task decomposition formulation, and validate our explanation through ablation experiments and visualizations.

%In this paper, we introduce a Cognition-driven XAI Framework (CDXF) that explains the internal mechanisms of Transformer-based models on integer arithmetic tasks. Specifically, we reformulate traditional arithmetic rules into a Transformer-friendly framework. Through three stages—Problem Analysis, Human Strategy Transfer, and Effect Verification—we leverage insights from cognitive science to diagnose accuracy issues and uncover their root causes. We then integrate human problem-solving strategies to enhance model performance, validate the efficacy of these improvements, and offer a detailed explanation of the model’s operation. 

\subsection{Data format}
Table \ref{table:data_format} summarizes the data format used in our experiments, including addition, multiplication, and division.
For multiplication, we validated the impact of reversing the answer digits, the formats of ordinal and reversed answer digits are marked as (O) and (R) in the table.
For multiplication and division, we also validated the impact of CoT, which is also illustrated in Table \ref{table:data_format}.

\subsection{Implementation Details}
\textbf{Training:} 
We conducted our experiments using a single-layer decoder-only transformer, which includes a Multi-Head Self-Attention (MHSA) layer and a Feed-Forward (FF) layer. The dimensions of MHSA and FF layers are 512 and 2048, respectively. For scaling experiments, we extend this to a multi-layer transformer with consistent dimensions. The model was trained with standard cross-entropy loss for next-token prediction, which is widely used in transformer training. The overall training process consists of 5000 iterations with a batch size of 64, using the Adam optimizer and a learning rate of 1e-4. 
%For $m\div u$ division, we set the dividend is a 6-digit number, and for $m\div m$ task, we set dividend and divisor are 8-digit and 3-digit number, respectively.
Each digit in the training data is independently sampled from a uniform distribution {0,1,...,9}. For the division task, we ensured that the first digit in divisor is non-zero in both $m\div u$ and $m\div m$ divisions. Table~\ref{table:data_format} summarizes the data format used in our experiments.

\textbf{Inference:} During inference, the input consists of dividend, division sign, divisor, and equal sign, \eg, ``$123456\times 7=$". The model generates each digit of the product answer in an autoregressive manner. The autoregressive process stops when generating the final digit. We test on 10k randomly sampled data that does not appear in the training set.

\section{Problem Analysis in Arithmetic Tasks}

\subsection{Addition}
\textbf{Subtask Decomposition:}
We begin our analysis with addition, the simplest arithmetic task. The addition task takes two $n$-digit numbers as input, producing an $n+1$-digit answer. For $5$-digit addition, there are 10 billion distinct questions and 200,000 possible answers. 
In addition, the calculation of each answer digit depends on specific digits in the addend and augend, as well as the carry from the previous position. Considering that the model calculates answer digits from the highest value digit, the calculation can be formulated as,

\begin{equation}
\begin{aligned}
A_i &= \lfloor(D_i+D_i')\\
    &+ (D_{i-1}+D_{i-1}')//10\\
    &+ ((D_{i-2}+D_{i-2}')//10+(D_{i-1}+D_{i-1}')\%10)//10\rfloor
\end{aligned}
\end{equation}
where $\lfloor\cdot\rfloor$ denotes floor operation.
To accurately calculate $A_i$, the Transformer needs to operate on three digit pairs, \eg, $(Di, D_i')$, $(D_{i-1}, D_{i-1}')$ and $(D_{i-2}, D_{i-2}')$.
Imitating the strategy humans adopt, we assert the transformer utilizes several basic subtasks to operate on digit pairs to complete the addition calculation.
\begin{itemize}
\item \textbf{Base Add (BA)}: BA calculates the sum of digit pair $(Di, D_i')$, which is the basic operation of the addition task.
\item \textbf{Make Carry (MC)}: MC evaluates whether there is a carry from the previous position.
\item \textbf{Make Sum 9 (MS9)}: MS9 determines whether the sum at the current position is 9, which is used to calculate consecutive carry operations.
\end{itemize}
Based on basic subtasks, the transformer chains multiple subtasks together to achieve complex tasks, \eg, \textbf{Use Carry (UC)} takes the carry from the previous column and adds it to the sum of the current digit pair, 
\textbf{Use Carry and Further Carry (UCFC)} uses carry from the previous column and further propagates a carry to the next column.

\begin{figure*}%[t]
\centering
\includegraphics[width=1\linewidth]{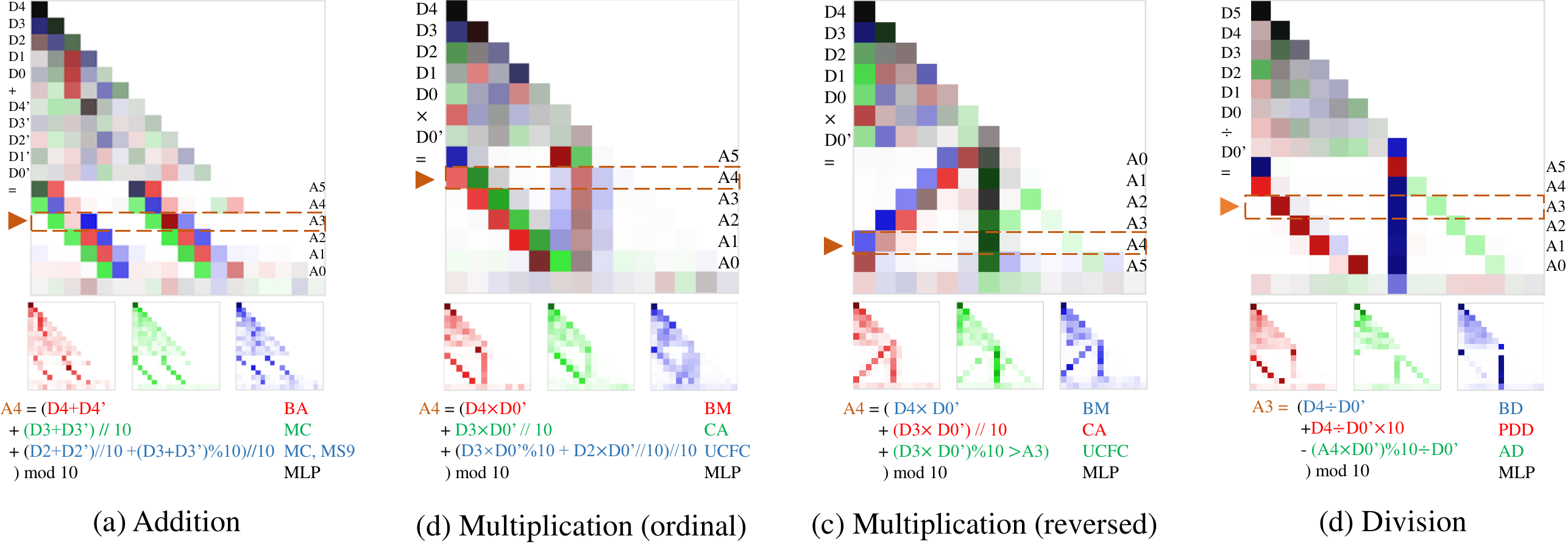}\\
% \vspace{-2mm}
\caption{Attention maps of arithmetic task: (a) addition, (b) multiplication(ordinal input), (c) multiplication(reversed input), and (d) division.}
% \vspace{-2mm}
\label{fig:attn}
\end{figure*}

\begin{figure*}
\centering
\includegraphics[width=0.85\linewidth]{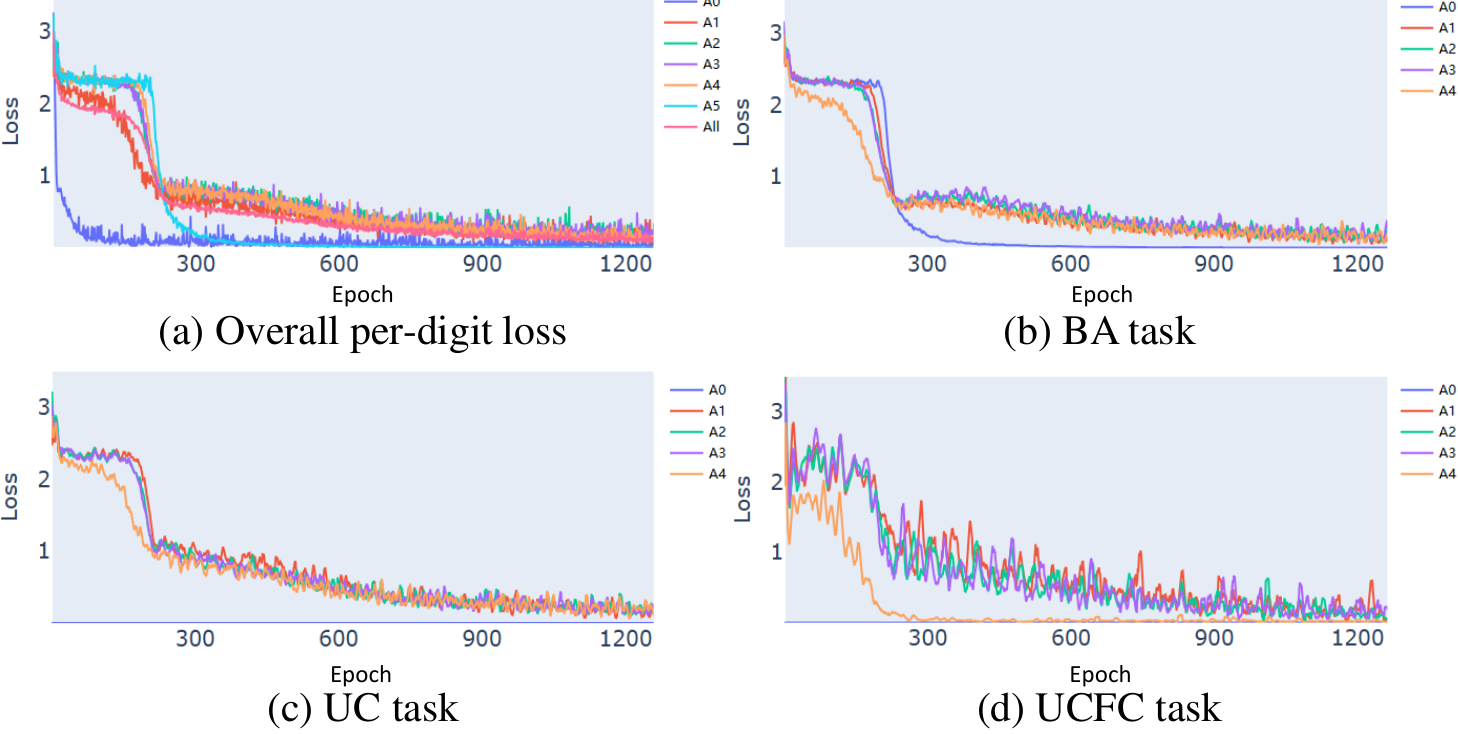}\\
\caption{Illustrations of (a) the overall per-digit loss curve, and (b-e) per-digit loss curve for each subtask on addition task.}
% \vspace{-4mm}
\label{fig:loss_add}
% \vspace{-2mm}
\end{figure*}

\textbf{Learning Analysis:}
Fig.\ref{fig:loss_add}(a) visualizes the per-digit and overall training loss for 5-digit addition task, with `A$n$'  representing the $n$-th digit and `All' indicating the overall loss. The per-digit loss curves show that the model learns each answer digit independently. The digit A0 converges much faster than the others because its calculation, which depends only on BA and does not require a carry from the previous column, is relatively straightforward. The loss for higher-order digits decreases more slowly, as they must account for the carry-over from lower digits.
We then categorized the training data into 3 non-overlapping subsets aligned to the BA, UC, and UCFC tasks, and visualizes the loss curve of each subtask in Fig.\ref{fig:loss_add}(b-d). As shown, BA and UC tasks show similar patterns, with the BA loss curve dropping more quickly because BA accuracy is needed before UC can be accurate. The curve of UCFC is much noisier than other tasks, as this task requires considering at least three digits, making it more complex than BA and UC.

\textbf{Visualization Analysis:}
Next, we validated the subtasks learned by Transformers through visualization analysis. We chose the number of attention heads that yielded the clearest separation of tasks in the attention patterns. Fig.~\ref{fig:attn}(a) shows that the model employs distinct attention heads for the BA and UC subtasks.
As shown, the heads attend to digits in the augend and addend sequentially from left to right. The calculation of answer digits A5 and A0 involves slightly different subtasks from other digits.
This attention pattern explains why the loss curve UCFC is noisier: each head concentrates on only three adjacent digits, which is insufficient for handling cascading UCFC scenarios. The experiment results in Table ~\ref{table:head} also show that Transformers with more than three heads effectively perform the addition task.

\subsection{Multiplication}

Unlike addition, multiplication involves a series of intermediate steps, particularly for large numbers. To streamline our analysis and interpretation, we start with multi-digit $\times$ unit-digit (m$\times$u) multiplication before extending our analysis to multi-digit $\times$ multi-digit (m$\times$m) multiplication.

\subsubsection{Unit-digit Multiplication}
\textbf{Subtask Decomposition:}
Based on the task decomposition formulation, the $m \times u$ multiplication can be formulated as,
\vspace{-1.5ex}
\begin{equation}
\begin{aligned}
A_i &= \lfloor(D_i\times D_0')\\
    &+ (D_{i-1}\times D_0')//10\\
    &+ (D_{i-1}\times D_0' \%10+D_{i-2}\times D_0'//\%10)//10\rfloor
\end{aligned}
\end{equation}
Imitating human strategy, we assert the transformer utilizes the following basic subtasks that operate on individual digit pairs,
\begin{itemize}
    \setlength{\itemsep}{0.0mm}
    \item \textbf{Base Multiply (BM)}: BM calculates the product of two single digits $D_i$ and $D_j'$ at each position.
    \item \textbf{Make Carry (MC)}: MC is responsible for calculating the carryover from the previous digit.
    %\item \textbf{Use Carry (UC)}: UC considers the carry from previous digit and add carry to the product at current position.
\end{itemize}
Similar to addition, complex functions can be achieved through combining simple tasks, \eg, \textbf{Use Carry (UC)} and \textbf{Use Carry and Further Carry (UCFC)}.% considers carry from previous column, and produces carry to next column.

\begin{table}
\begin{center}
\caption{Accuracy (\%) of Transformers on arithmetic tasks with different attention heads, Mul (O) and Mul (R) denote multiplication task with ordinal and reversed answer digits.}
\setlength{\tabcolsep}{12 pt}
\small
\begin{tabular}{c|cccc}
\toprule
\#Heads &Add  &Mul (O) & Mul (R) &Div  \\\midrule
1       &89.1 &66.2  &83.6 &31.7 \\\midrule
2       &98.7 &85.3  &96.3 &86.2 \\\midrule
3       &99.9 &89.8  &99.8 &99.8 \\\midrule
4       &100  &90.1  &100 &99.9 \\\midrule
5       &100  &91.1  &100  &100 \\\midrule
6       &100  &91.2  &99.9 &100 \\\midrule
%GPT2      &13.1  &18.5  \\\bottomrule
\end{tabular}
\label{table:head}
% \vspace{-2mm}
\end{center}
\end{table}

\begin{figure*}
\centering
\includegraphics[width=0.85\linewidth]{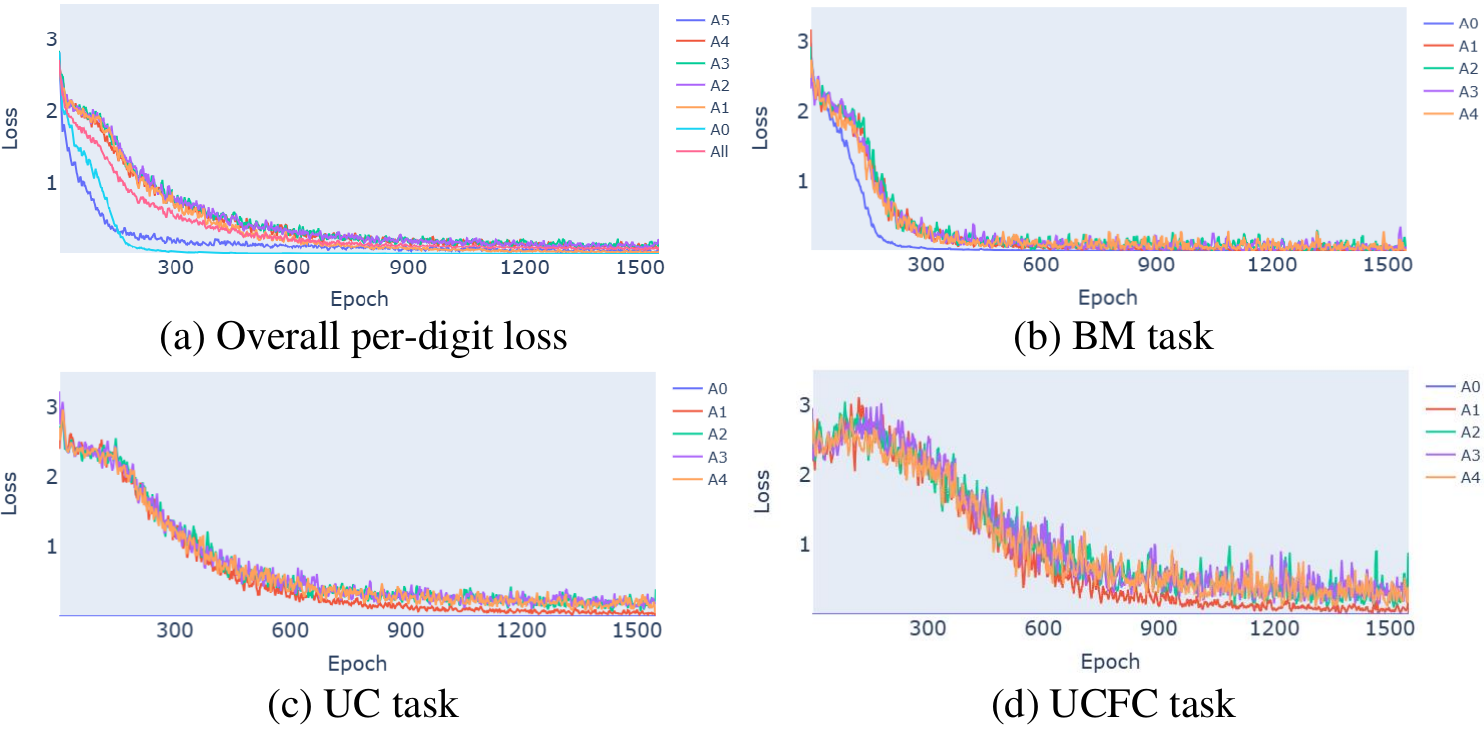}\\
% \vspace{-2mm}
\caption{Illustrations of (a) the overall per-digit loss curve, and (b-e) per-digit loss curve for each subtask on multiplication task.}
\label{fig:loss_mul}
% \vspace{-2mm}
\end{figure*}

\textbf{Learning Analysis:}
Fig.~\ref{fig:loss_mul}(a) shows the per-digit and overall training loss for 5-digit $m \times u$ multiplication, with `A$n$' representing the $n$-th digit and `All' indicating the overall loss. Similarly, the curves reveal that Transformers learn each answer digit semi-independently. A0 and A5 are learned faster and with less noise, as A0 requires no carry, and A5 only considers the carry. In contrast, A1-A4 show similar decline patterns, as each depends on both the carry from the previous column and the current product, contributing to the carry for the next column.
We then analyze each subtask in Fig.~\ref{fig:loss_mul}(b-d), categorizing the training data into non-overlapping subsets for each subtask and showing per-digit loss curves. A notable pattern emerges: initially, all tasks show high loss, but as training progresses, the loss curves of each task drop with a `time lag'. The BM curve drops fastest as expected, since BM underpins all other tasks.
The UC loss is the second-fastest to decline, initially plateauing and indicating its dependency on BM. This suggests that the BM starts with simpler, no-carry scenarios and only later adapts to more complex UC cases, which explains the second pronounced drop in BM’s loss curve. 
UCFC curve decreases last with significant noise, because it spans three digits and is inherently more complex than the other tasks.

% \section{Verification of learned subtasks by attention head}

\textbf{Verification of learned subtasks by attention head:}

We validated the learned subtasks by each attention head of the reversed transformer.
For each head, we used an ablation intervention technique~\cite{quirke2023understanding} that overrode its output with the mean value of the whole dataset and computed the average loss on specific samples related to the target $BM$ task. 

The experimental results on the BM task are shown in table~\ref{table:ablate1}. 
As shown, ablating head 0 and head 1 has a minor impact on the loss of the BD task, while ablating head 2 causes a significant loss increase.
We hence conclude that head 2 is the key for the calculation of the BM task. Similar conclusions can also be observed on other heads in Table~\ref{table:ablate2} and ~\ref{table:ablate3}. This indicates that each attention head has a well-defined role, and they collaborate to accomplish the division task.

\begin{table}%[h]
\begin{center}
\caption{The influence of ablating each attention on \textbf{BM} task. }
\setlength{\tabcolsep}{10 pt}
\footnotesize
\begin{tabular}{c|c|l}
\toprule
Ablated Head  &Avg loss &Conclusion  \\\midrule
-        &0.000 &-  \\\midrule
0        &0.315 &Minor impact.  \\\midrule
1        &0.632 &Minor impact.  \\\midrule
2        &5.552 &Large impact, head 2 is key for BM. \\\midrule
\end{tabular}
\label{table:ablate1}
% \vspace{-3mm}
\end{center}
\end{table}

\begin{table}%[h]
\begin{center}
\caption{The influence of ablating each attention on \textbf{CA} task. }
\setlength{\tabcolsep}{10 pt}
\footnotesize
\begin{tabular}{c|c|l}
\toprule
Ablated Head  &Avg loss &Conclusion  \\\midrule
-        &0.001 &-  \\\midrule
0        &7.846 &Large impact, head 0 is key for CA.  \\\midrule
1        &0.979 &Minor impact.  \\\midrule
2        &1.258 &Minor impact. \\\midrule
\end{tabular}
\label{table:ablate2}
% \vspace{-3mm}
\end{center}
\end{table}

\begin{table}%[h]
\begin{center}
\caption{The influence of ablating each attention on \textbf{UCFC} task. }
\setlength{\tabcolsep}{10 pt}
\footnotesize
\begin{tabular}{c|c|l}
\toprule
Ablated Head  &Avg loss &Conclusion  \\\midrule
-        &0.002 &-  \\\midrule
0        &0.784 &Minor impact.  \\\midrule
1        &6.161 &Large impact, head 1 is key for UCFC.  \\\midrule
2        &1.116 &Minor impact. \\\midrule
\end{tabular}
\label{table:ablate3}
% \vspace{-3mm}
\end{center}
\end{table}

\begin{figure*}
\centering
\includegraphics[width=0.99\linewidth]{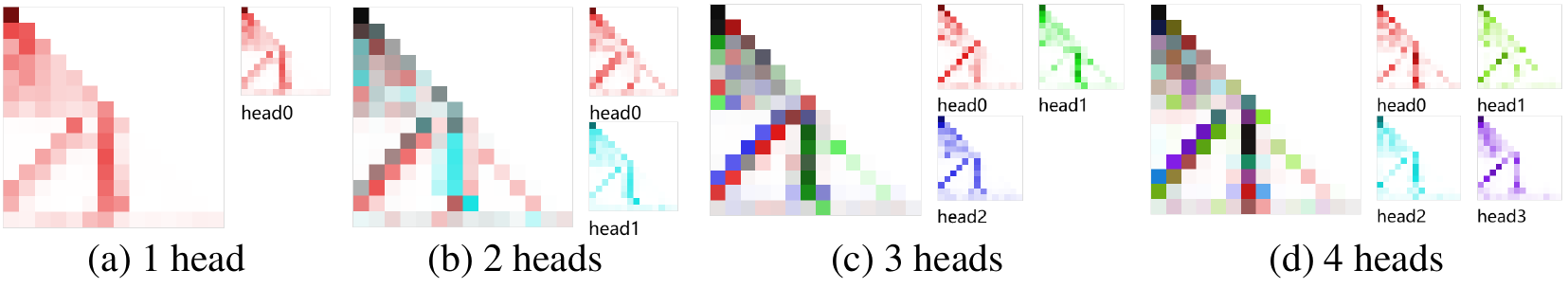}\\
\caption{The attention map of reversed transformer with different attention heads.}
\label{fig:attnmap_head}
\end{figure*}

\textbf{Visualization Analysis:}
In this section, we analyze the model's prediction behavior through visualization. Inspired by humans performing multiplication, which typically starts from lower digits~\cite{lee2023teaching}, we also visualize the attention patterns with reversed answer digits. In Fig.\ref{fig:attn}(b)(c), we visualize the attention patterns of ordinal and reversed transformers, the data formats are provided in Table~\ref{table:data_format}. 
In Fig.\ref{fig:attn}(b), the heads attend to digits in the multiplicand sequentially from left to right. Three attention heads have a 1-token offset, with each head handling specific subtasks, \eg, the red head computes BM on D4 and D0'. The attention map also explains the low accuracy of the ordinal Transformer: it can only focus on three continuous tokens in the multiplicand, which is insufficient for cascaded UCFC tasks that require at least four tokens.
In Fig.~\ref{fig:attn}(b), each attention head is also responsible for specific subtasks. This attention pattern explains why the reversed Transformer performs better: it can leverage previously generated answer digits to compute the current digit, enabling it to better handle cascaded UCFC cases. The visualizations of transformers with different heads are shown in Fig.\ref{fig:attnmap_head}. As shown, when the head number is less than 3, multiple tasks are packed into a single head, hence causing the low accuracy. Conversely, when there are more than 3 heads, multiple heads end up performing the same subtask. 
The model with 3 heads demonstrates the clearest task separation in attention patterns, it has separate attention heads for BA, CA, and UCFC tasks.

Table ~\ref{table:head} shows the performance of Transformers with different heads. In the table, the reversed Transformer consistently achieves superior accuracy with different numbers of attention heads, and performance improvement tends to saturate with more than 3 attention heads, this indicates 3 heads are sufficient to complete multiplication.
We also validated the subtasks learned by each head through ablation experiments. The ablation experiments confirm that each attention head has a distinct, well-defined function.
%We validated the subtasks learned by each head through ablation experiments in the Appendix-C.1. The ablation experiments confirm that each attention head has a distinct, well-defined function.

\begin{figure}
\centering
\includegraphics[width=0.9\linewidth]{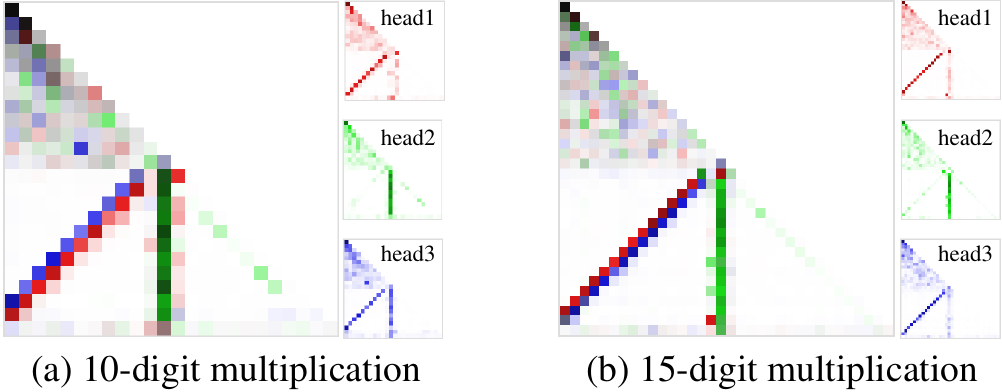}\\
\caption{The attention map of transformer on 10-digit and 15-digit multiplication.}
\label{fig:attnmap_1015}
\end{figure}

\textbf{Algorithm reuse:}
We explored whether the subtasks are learned by similar Transformer-based models. We first train the same model on the 5-digit multiplication task with different random seeds or optimizers. The resulting models show similar behavior to the previous one. We then train Transformers on 10-digit and 15-digit multiplication, and visualize the attention maps in Fig.\ref{fig:attnmap_1015}. The new models also use BM, UC and UCFC subtasks to complete the multiplication calculation. This analysis suggests that the Transformer architecture, when trained on the multiplication task, converges to a consistent algorithm for performing multiplication, indicating a robust algorithmic solution emerges from the Transformer’s architecture and training.

\begin{table}
\begin{center}
\caption{Overall and per-digit accuracy(\%) with various multipliers. }
\setlength{\tabcolsep}{3 pt}
\small
\begin{tabular}{c|c|cccccccccc}
\toprule
Format &overall     &A9&A8&A7&A6&A5&A4&A3&A2&A1&A0    \\\midrule
$0000d$  &85.3&100 &100 &100 &100 &98.4&95.4&94.4&94.9&99.6&100    \\\midrule
$d000d$  &16.8&98.6&96.9&95.2&85.1&40.2&28.6&96.0&96.4&99.5&100    \\\midrule
$000dd$ &5.2 &100 &100 &100 &93.8&45.2&21.6&19.6&20.0&32.0&100    \\\midrule
$00ddd$ &0.2 &100 &100 &94.3&43.1&13.4&11.3&10.8&12.4&31.6&100    \\\midrule 
$0dddd$ &0   &100 &95.2&44.4&11.8&10.0&10.4&9.3 &12.4&31.4&100    \\\midrule 
$ddddd$ &0   &91.3&39.9&13.4&10.5&9.9 &9.5 &10.5&12.3&32.6&100    \\\midrule 
%\bottomrule
\end{tabular}
\label{table:digitacc_mul}
% \vspace{-3mm}
\end{center}
\end{table}

\subsubsection{Multi-digit Multiplication}
This section investigates the $m \times m$ multiplication task, where the vanilla Transformer exhibits relatively low performance. We analyze the reasons for the Transformer's limitations, and explore potential solutions by examining internal mechanisms inspired by human learning processes.

\begin{figure}
\centering
\includegraphics[width=0.80\linewidth]{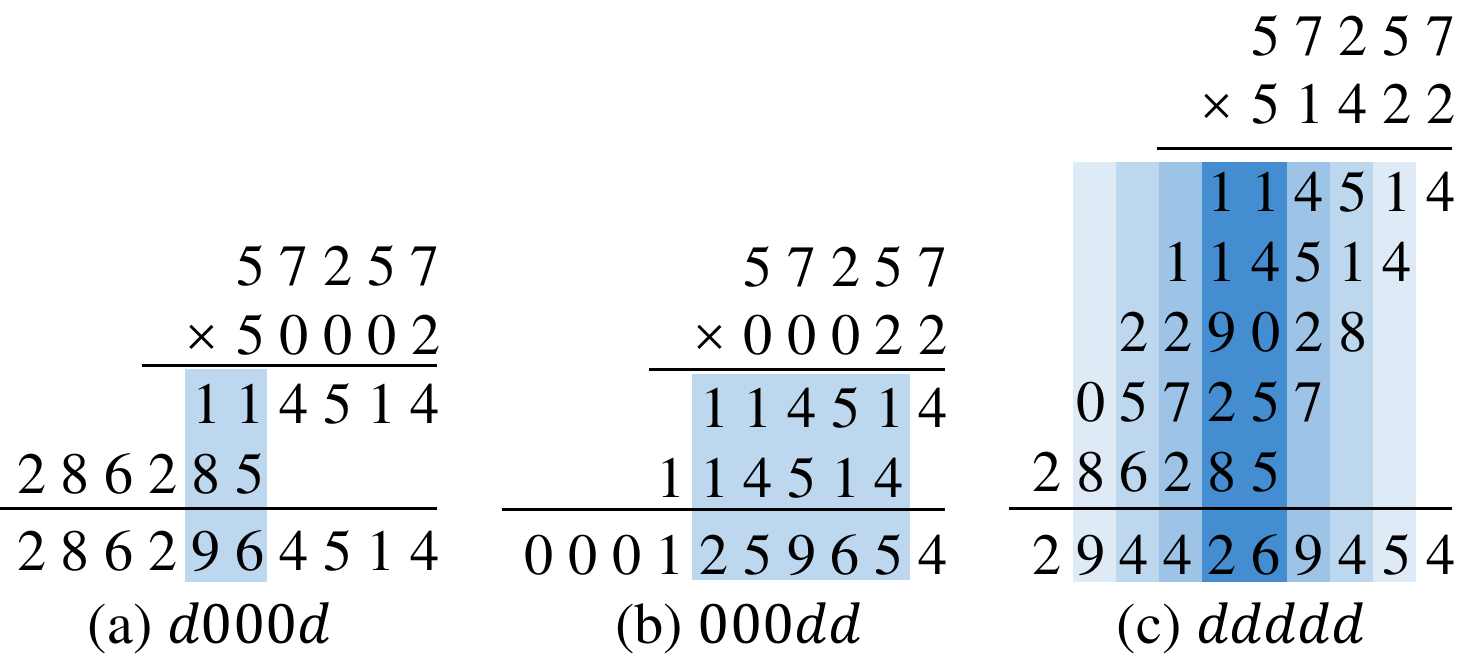}\\
\caption{The overlap of per-digit products with different multiplier formats, with darker colors indicating more overlapping digits.}
\label{fig:overlap}
\end{figure}

\textbf{Human strategy transfer:}
We first examine the per-digit accuracy of ordinal and reversed Transformers with various multipliers, with results summarized in Table~\ref{table:digitacc_mul}. The accuracy drop correlates with the overlap between intermediate products of the multiplicand and each multiplier digit. For example, for the $ddddd \times ddddd$ format in Fig.\ref{fig:overlap}(c), digits A4 and A5 require consideration of all intermediate products, resulting in the lowest accuracy. This suggests that the vanilla Transformer struggles with the complex intermediate steps in $m \times m$ multiplication.

Humans employ specific strategies when performing arithmetic tasks including computational strategy, chain of thought reasoning, and cognitive resource allocation. These strategies corresponding to change calculation order (reverse answer digit), Chain of Thought (CoT) prompt and increasing model depth. In following part, we verify the effectiveness of these strategies, and provide comprehensive interpretation. 
We first validated the effectiveness of these strategies in Table~\ref{table:refine}.
Reversing answer digits is effective in the $m\times u$ task, but does not bring obvious improvement when used alone in the $m\times m$ task. This is because, even with digit reversal, the $m\times m$ task still involves complex intermediate steps that are difficult for a single-layer model to handle. CoT decomposes the complex $m\times m$ multiplication into $m\times u$ multiplication, reducing the prediction difficulty of each step. Similarly, increasing the model depth can increase model's capacity, allowing it to handle more intermediate steps. These strategies correspond to the way humans perform multiplication, and their combination achieves best accuracy.

\begin{figure*}
\centering
\includegraphics[width=0.99\linewidth]{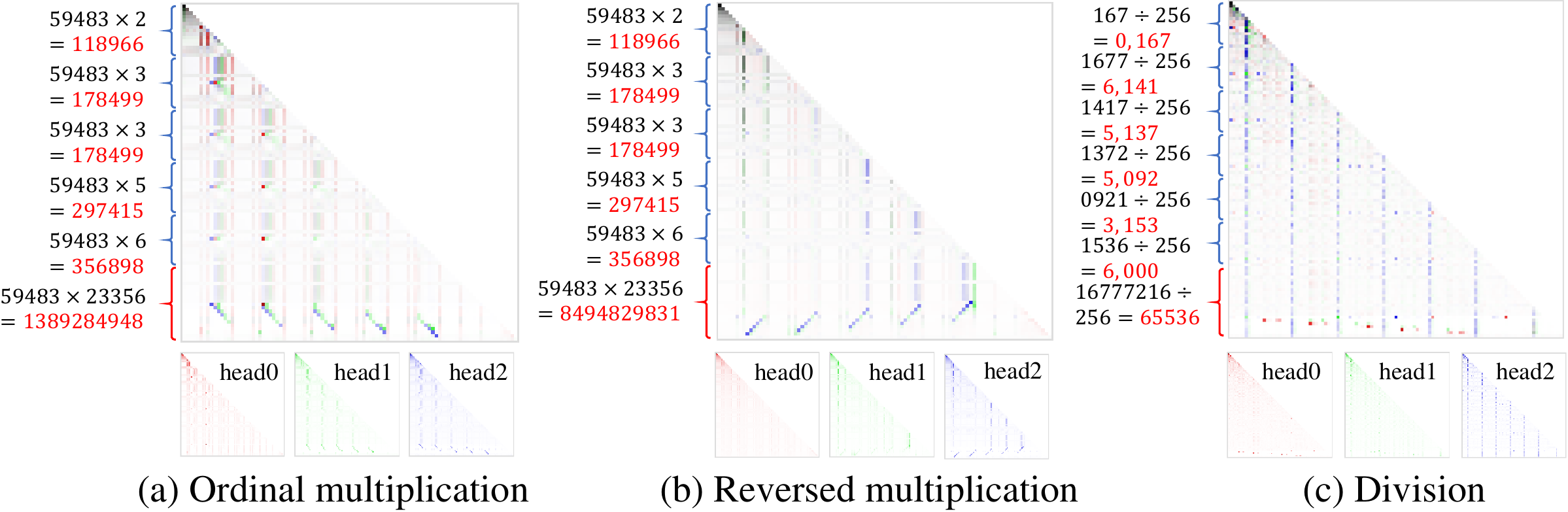}\\
\caption{The attention map of transformer with CoT input, (a) ordinal transformer on multiplication task, (b) reversed transformer on multiplication task, and (c) ordinal transformer on division task. The characters that the transformer needs to predict are highlighted in red.}
\label{fig:attnmap-cot}
\end{figure*}

\textbf{Visualization Verification:}
We visualize the attention maps of ordinal and reversed Transformers with CoT input in Fig.~\ref{fig:attnmap-cot}(a) and (b). CoT decomposes the $m \times m$ task into simpler $m \times u$ multiplications and additions, reducing calculation complexity. 
Since the addition task is also easier when computed from lower-order digits, the reversed Transformer also outperforms the ordinal transformer with CoT inputs.

\subsection{Division}

Different from addition and multiplication, the calculation order of division is reversed. In this section, we also conduct analysis starting with the simpler $m \div u$ division.

\subsubsection{4.3.1 Unit-digit Divisor Division}
\textbf{Subtask Decomposition:}
We first decompose the division as pairwise digit operation based on our task decomposition formulation,
\vspace{-1ex}
\begin{equation}
    A_i = \lfloor\frac{D_i}{D_0'} + \frac{D_{i+1}}{D_0'}\times 10-\frac{(A_{i+1}\times D_0')\%10}{D_0'}\times 10\rfloor
\end{equation}
where $\lfloor\cdot\rfloor$ denotes floor operation.
To accurately calculate $A_i$, the model needs to operate on 3 digit pairs, \eg, $(Di, D_0')$, $(D_{i+1}, D_0')$ and $(A_{i+1}, D_0')$.
We assert the transformer utilizes several basic subtasks to operate on these digit pairs.

\begin{itemize}
\item \textbf{Base Division (BD)}: BD operates on digit pair $(Di, D_0')$ to calculates the base quotient of two single digits. 
\item \textbf{Previous Digit Division (PDD)}: PDD calculate the quotient of the previous digit $D_{i+1}$ and divisor $D_0'$, which is subsequently used for remainder division calculations.
\item \textbf{Answer Division (AD)}: AD is response for the calculation of the portion occupied during the calculation of the previous digit. 
\end{itemize}

\begin{table}
\caption{Effectiveness of refinements on multiplication and division.}
% \vspace{-3mm}
\setlength{\tabcolsep}{10 pt}
% \small
\begin{center}
\begin{tabular}{ccc|c|cc}
\hline
\multirow{2}{*}{Reserve} &\multirow{2}{*}{Depth} &\multirow{2}{*}{CoT} &\multicolumn{2}{|c}{Accuracy}\\\cline{4-5}
 & & &Multiply &Division\\
\hline
&  &  &0.0 &1.2\\ \hline
\checkmark&  &           &0.0  &0.0\\
&\checkmark  &           &79.1 &65.1\\ 
&  &\checkmark           &80.2 &83.9\\\hline
\checkmark &\checkmark & &99.9 &30.1\\
\checkmark & &\checkmark &99.6 &-\\
&\checkmark &\checkmark  &99.3 &\textbf{100}\\\hline
\checkmark&\checkmark&\checkmark &\textbf{100} &-\\
\hline
\end{tabular}
\end{center}
\label{table:refine}
% \vspace{-3mm}
\end{table}

\begin{figure*}
\centering
\includegraphics[width=0.85\linewidth]{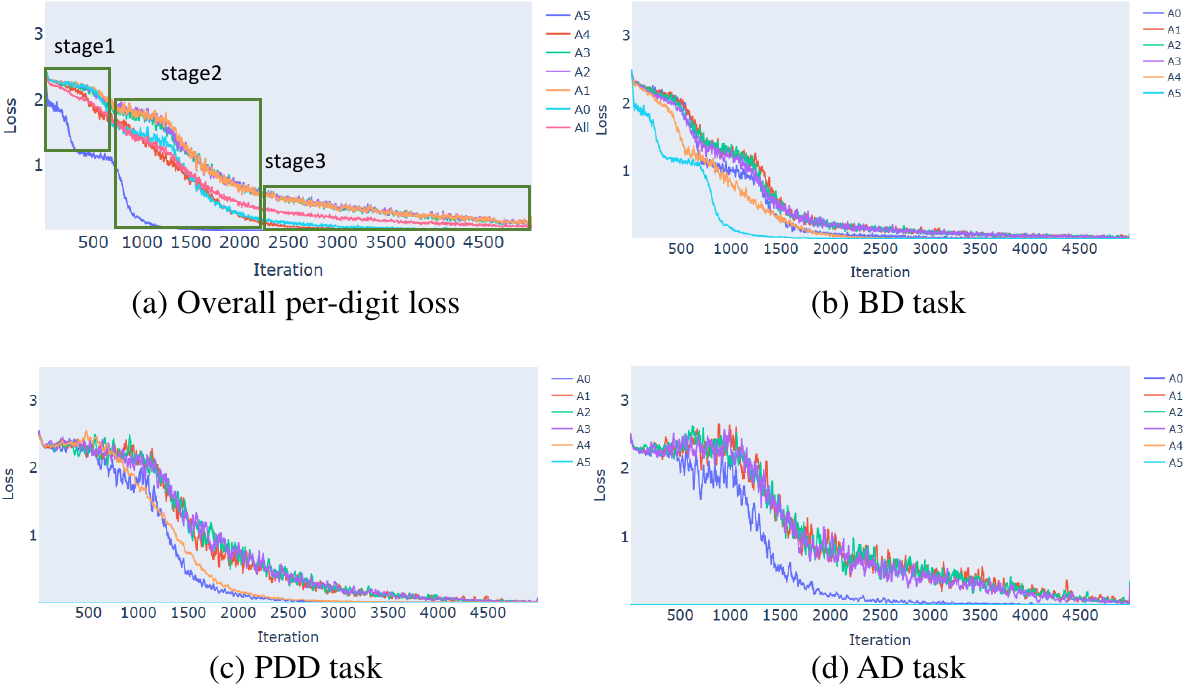}\\
% \vspace{-2mm}
\caption{Illustrations of (a) the overall training loss curve and (b-d) per-digit loss curves for division task.}
% \vspace{-2mm}
\label{fig:loss_div}
\end{figure*}

\textbf{Learning Analysis:}
We first examine the overall training behavior of the Transformer on division, as shown in Fig.~\ref{fig:loss_div}(a). `A$n$' represents the $n$-th digit in answer digit, and `$\operatorname{All}$' denotes the overall loss curve for all digits. The training process can be divided into three stages. The first two stages correspond to rapid loss reduction, while the third stage shows a slower decrease as all digits are well learned, resulting in smooth and low loss. The fast loss reduction in stages 1 and 2 reflects the Transformer’s `grokking' of the rules of division.
The loss curves in Fig.~\ref{fig:loss_div}(a) also show that Transformers learn each answer digit semi-independently. A5 is learned much more quickly than other digits, as it does not need to consider any remainder from the previous digit. The loss curves for A4 to A1 follow a similar pattern, but their decrease shifts progressively later, as their calculation depends on the remainder from the previous digit. A0 exhibits a distinct learning pattern, decreasing faster than A4 to A1. This is because the model can `guess' the A0 based on divisor and the last digit of dividend, \eg, when $D0=4$ and $D'=6$, $A0$ must be either 4 or 9, making its calculation simpler than the others.

We then analyze the Transformer's learning behavior for each subtask by categorizing the training data into non-overlapping subsets aligned with each subtask and illustrating the per-digit loss curves in Fig.\ref{fig:loss_div}(b-d). Focusing on the BD task in Fig.\ref{fig:loss_div}(b), its loss decreases much faster than the others as expected, since it is the simplest task. This rapid loss decline aligns with stage 1 in Fig.~\ref{fig:loss_div}(a). The losses for the PDD and AD tasks decrease after BD, as they rely on the results from the previous digits. This analysis reveals the Transformer's learning behavior for division tasks.

\textbf{Visualization Analysis:}
In this section, we investigate how Transformers perform division via visualization. Fig.~\ref{fig:attn}(d) visualizes the attention map, which shows that each attention head focuses on a distinct digit pair to perform specific subtasks. For example, the blue head handles the BD task based on digit $D_i$ in dividend and $D_0'$ in divisor. The outputs of these attention heads are combined in the subsequent MLP layer to generate the final answer digit. %We further validated the subtasks learned by each head through ablation experiments, with results summarized in the supplementary materials. The experiments confirm that each attention head has a clear role and collaborates to complete the division task.

% \begin{table}
% \begin{center}
% \caption{The influence of ablating each attention on \textbf{BD} task. Head 2 plays a crucial role in the calculation of \textbf{BD} task.}
% \setlength{\tabcolsep}{2 pt}
% \small
% \begin{tabular}{c|c|l}
% \toprule
% Ablated Head  &Avg loss &Conclusion  \\\midrule
% -        &0.001 &-  \\\midrule
% 0        &0.453 &Minor impact.  \\\midrule
% 1        &0.314 &Minor impact.  \\\midrule
% 2        &4.456 &Large impact, head 2 is the key of BD. \\\midrule
% %GPT2      &13.1  &18.5  \\\bottomrule
% \end{tabular}
% \label{table:ablate}
% \end{center}
% \end{table}

%\subsubsection{Algorithm Reuse}
%We investigated whether the sub-tasks learned by aforementioned transformer exhibit generalization capabilities. We attempted to train the same model with different random seeds, or train different models to perform division tasks with varying dividend digit lengths.

%The experimental results show that these newly trained models perform the subtasks in the same manner as the previous model. This suggests that the transformer trained on division task converges to a consistent algorithm. This algorithm utilizes the self-attention mechanism to identify and execute the required subtasks in parallel. Despite variations in random initialization and output format, the transformers demonstrate similar internal behavior and capabilities, indicating that a robust algorithmic solution emerges from both the transformer architecture and the training process.

\subsubsection{Multi-digit Division Analysis}

The vanilla Transformer also performs poorly on $m\div m$ division tasks. This section explores the reasons for the accuracy drop and provides an explanation of the internal mechanisms of potential solutions inspired by human learning processes.

\textbf{Human strategy transfer:}
We first analyze per-digit accuracy on $m \div m$ division with various dividends formats in Table~\ref{table:digitacc} (\eg, `$00000ddd$' represents an 8-digit number with the first 5 digits fixed as zero). When the dividend is $dddddddd$, the model achieves very low accuracy (1.2\%). Notably, accuracy decreases sequentially from left to right as we move to non-zero dividend digits, reflecting the accumulation of errors in remainder calculation. 
This indicates that the vanilla single-layer model struggles to handle the intermediate remainder results. However, the accuracy for the units digit (\eg, $A0$) remains consistently high across different dividend formats. 
The reason is similar to $A0$ in $m \div u$ division: the Transformer can `guess' the units digit based on the divisor and dividend, avoiding error accumulation.

\begin{table}
\begin{center}
\caption{Overall and per-digit accuracy (\%) with various dividend. }
% \vspace{-2mm}
\setlength{\tabcolsep}{6 pt}
\small
\begin{tabular}{c|c|ccccccc}
\toprule
dividend &overall &A5&A4&A3&A2&A1&A0       \\\midrule
$00000ddd$     &57.2 &100 &100 &100 &100 &57.2&95.7   \\\midrule
$0000dddd$     &21.6 &100 &100 &100 &100 &22.0&90.6    \\\midrule
$000ddddd$     &19.1 &100 &100 &100 &87.4&22.4&90.4   \\\midrule
$ddddd000$     &9.7  &97.2&62.1&69.7&15.2&100 &100   \\\midrule
$dddddddd$      &1.2  &98.7&89.1&45.8&12.7&10.5&90.5    \\\midrule 
%\bottomrule
\end{tabular}
\label{table:digitacc}
% \vspace{-2mm}
\end{center}
\end{table}

We then studied the strategies of human experience, include reversing answer digits, increasing model depth and use Chain of Thought (CoT), and analyzed the underlying principles behind these strategies. 
We first validated the effectiveness of these strategies in Table~\ref{table:refine}.
Reversing the digits is effective for multiplication, but it actually reduces performance for division. This is because division calculations start from the higher-order digits, which aligns with human experience. Increasing the model depth can improve the accuracy of division calculations. This is because deeper models have greater capacity to handle intermediate results, leading to more accurate division calculations. CoT decomposes complex tasks, effectively reducing the model's prediction difficulty. The effects of these strategies align with human experience.

\textbf{Visualization Verification:}
We visualized the attention map of Transformers with CoT input in Fig.~\ref{fig:attnmap-cot}(c). As shown, when performing $m\div m$ division, the Transformer not only focuses on dividend and divisor, but also attends to the previous intermediate results. This decomposes the complex division task into multiple simpler ones, reducing prediction difficulty.

% \begin{table}
% \caption{Accuracy (\%) of ordinal and reversed transformer with different depth.}
% \setlength{\tabcolsep}{15 pt}
% \small
% \begin{center}
% \begin{tabular}{c|cc}
% \toprule
% \multirow{2}{*}{\# Layers} &\multicolumn{2}{|c}{Format}  \\
% \cmidrule{2-3}
%  &Ordinary &CoT     \\\midrule
% 1         &1.2  &83.9   \\\midrule
% 2         &4.2  &84.7   \\\midrule
% 4         &15.3 &89.8   \\\midrule
% 8         &47.6 &99.8   \\\midrule
% 12        &65.1 &99.9   \\\midrule
% \end{tabular}
% \end{center}
% \label{table:refine}
% \end{table}

%we then investigated the influence of model depth. Table~\ref{table:reversed_acc} shows the accuracy of transformer models with different layers. As shown, increasing model depth can enhance its capacity, allowing it to store more intermediate results, thereby achieving higher accuracy. Reversed-order computation requires fewer intermediate results, therefore the model with fewer layers is sufficient to achieve higher accuracy. While the ordinal-order computation requires more intermediate steps because the model cannot utilize the previously generated answer digits, necessitating a deeper model to store these intermediate results.
% \vspace{-1ex}
\subsection{Analysis of LLMs in Arithmetic}
Based on analysis in this paper, we analyze the possible reasons for the suboptimal performance of current LLMs on arithmetic tasks. 

\begin{itemize}
\item Model capacity: although LLMs have a large capacity, due to the scarcity of arithmetic data in the internet-collected training data, only a small fraction of neurons are `specialized' in arithmetic, hence the capacity is insufficient for arithmetic calculation.
\item Data format: in internet-collected data, the arithmetic data are usually presented in ordinal order. However, arithmetic tasks have different calculation orders, \eg, the multiplication and division tasks follow entirely different calculation orders.
\item Output diversity: NLP tasks allow more diverse outcomes hence the LLM are encouraged diversified outputs, while arithmetic calculations have definite results.
%\item Tokenizer: in some LLMs' tokenizer, multiple digits are packaged as a single token , \eg, in GPT-4, the input arithmetic formula ``$12345\times 857$" is divided into ``123", ``45", ``$\times$" and ``857". This tokenizer further increases the calculations difficulty of arithmetic.
\end{itemize}

% \vspace{-2ex}

\section{Conclusion}

This paper presents a comprehensive problem analysis to inspect and explain Transformer's implementation on arithmetic tasks. Our findings demonstrated that the Transformer decomposes the arithmetic task into multiple parallel streams and combines the partial results to yield the final outcome. Through comprehensive learning and visualization analysis, we explain the underlying mechanisms of the Transformer's inferior performance on arithmetic tasks. Furthermore, we enhanced the model’s accuracy and interpretability through transferring human cognitive skills, including computational strategy, Chain-of-Thought(CoT) reasoning, and cognitive capacity empowerment, suggesting that Transformer-based models may exhibit learning patterns analogous to human problem-solving strategies in arithmetic. Our approaches contribute to the broader fields of model understanding and interpretability. It paves the way for enhancing the interpretability of more complex tasks and larger Transformer-based models.

\end{document}